\documentclass[conference]{IEEEtran}
\IEEEoverridecommandlockouts
\usepackage{cite}
\usepackage{amsmath,amssymb,amsfonts}
\usepackage{algorithmic}
\usepackage{graphicx}
\usepackage{textcomp}
\usepackage{xcolor}
\usepackage{algorithm}
\usepackage{multirow}
\usepackage{makecell}
\def\BibTeX{{\rm B\kern-.05em{\sc i\kern-.025em b}\kern-.08em
    T\kern-.1667em\lower.7ex\hbox{E}\kern-.125emX}}
\begin{document}

\makeatletter
\newcommand{\linebreakand}{%
  \end{@IEEEauthorhalign}
  \hfill\mbox{}\par
  \mbox{}\hfill\begin{@IEEEauthorhalign}
}
\makeatother

\title{$H^{2}R$: Hierarchical Hindsight Reflection for Multi-Task LLM Agents
}

% \author{\IEEEauthorblockN{Anonymous Authors}}
% \author{
% \IEEEauthorblockN{1\textsuperscript{st} Shicheng Ye}
% \IEEEauthorblockA{\textit{Sun Yat-sen University} \\
% yeschch@mail2.sysu.edu.cn}
% \and
% \IEEEauthorblockN{2\textsuperscript{nd} Chao Yu}
% \IEEEauthorblockA{\textit{Sun Yat-sen University} \\
% yuchao3@mail.sysu.edu.cn}
% \and
% \IEEEauthorblockN{3\textsuperscript{rd} Kaiqiang Ke}
% \IEEEauthorblockA{\textit{Sun Yat-sen University} \\
% kekq@mail2.sysu.edu.cn}
% \linebreakand
% \IEEEauthorblockN{4\textsuperscript{th} Chengdong Xu}
% \IEEEauthorblockA{\textit{Sun Yat-sen University} \\
% xuchd6@mail2.sysu.edu.cn}
% \and
% \IEEEauthorblockN{5\textsuperscript{th} Yinqi Wei}
% \IEEEauthorblockA{\textit{The University of Sydney} \\
% weiy0041@e.ntu.edu.sg}
% }

\author{
% \IEEEauthorblockN{1\textsuperscript{st} Given Name Surname}
\IEEEauthorblockN{1\textsuperscript{st} Shicheng Ye}
\IEEEauthorblockA{\textit{Sun Yat-sen University} \\
yeschch@mail2.sysu.edu.cn}
\and
\IEEEauthorblockN{2\textsuperscript{nd} Chao Yu}
\IEEEauthorblockA{\textit{Sun Yat-sen University} \\
yuchao3@mail.sysu.edu.cn}
\and
\IEEEauthorblockN{3\textsuperscript{rd} Kaiqiang Ke}
\IEEEauthorblockA{\textit{Sun Yat-sen University} \\
kekq@mail2.sysu.edu.cn}
% \and
\linebreakand
\IEEEauthorblockN{4\textsuperscript{th} Chengdong Xu}
\IEEEauthorblockA{\textit{Sun Yat-sen University} \\
xuchd6@mail2.sysu.edu.cn}
\and
\IEEEauthorblockN{5\textsuperscript{th} Yinqi Wei}
\IEEEauthorblockA{\textit{The University of Sydney} \\
weiy0041@e.ntu.edu.sg}
}

\maketitle

\begin{abstract}
Large language model (LLM)-based agents have shown strong potential in multi-task scenarios, owing to their ability to transfer knowledge across diverse tasks. 
However, existing approaches often treat prior experiences and knowledge as monolithic units, leading to inefficient and coarse-grained knowledge transfer. 
In this work, we propose a novel hierarchical memory architecture that enables fine-grained knowledge transfer by decoupling high-level planning memory from low-level execution memory. To construct and refine these hierarchical memories, we introduce Hierarchical Hindsight Reflection ($H^{2}R$), a mechanism that distills reusable and hierarchical knowledge from past agent–environment interactions. 
At test time, $H^{2}R$ performs retrievals of high-level and low-level memories separately, allowing LLM-based agents to efficiently access and utilize task-relevant knowledge for new tasks.
Experimental results across two benchmarks demonstrate that $H^{2}R$ can improve generalization and decision-making performance, outperforming prior baselines such as Expel. 
% This work provides a scalable and effective solution for sustainable cross-task knowledge scaffolding in multi-task LLM agents.
\end{abstract}


\section{Introduction}
\label{sec-1}
% Multi-task learning~\cite{caruana1997multi-task} is widely regarded as a key milestone toward general artificial intelligence, as it requires agents to transfer knowledge effectively across diverse tasks. To achieve this, agents need the ability to generalize and selectively reuse prior experience and knowledge. LLM-based agents are particularly promising because their generalization and in-context learning capabilities allow knowledge and skills acquired from previous tasks to be adapted to novel contexts, which improves performance and adaptability in multi-task scenarios. 
Multi-task learning~\cite{caruana1997multitask} is a key step toward general artificial intelligence, as it requires agents to handle a variety of tasks with distinct goals and requirements.
Due to the strong generalization and in-context learning abilities of large language models (LLMs), LLM-based agents are particularly well suited to multi-task settings by adapting knowledge and skills acquired from previous tasks to novel situations~\cite{huang2024understanding}.

A common paradigm for LLM-based agents to solve multi-tasks is to construct a memory repository storing task-solving insights extracted from past interactions, such as comprehension of environmental dynamics~\cite{chari2025mindstores}, improvement of task execution plans~\cite{zhao2024expel,fu2024autoguide,chen2024automanual}, and correction of errors~\cite{shinn2023reflexion}. 
When dealing with new tasks, the agents can selectively retrieve the most relevant memories from this repository to inform decision making, thereby improving performance in multi-task scenarios. For example, an agent in a household environment that has learned to ``\textit{clean a pan and place it on the countertop}'' could reuse the knowledge from prior experiences to execute a new task like ``\textit{cooling lettuce and placing it on the countertop}'' more effectively.

% However, existing approaches often treat episodic experiences and insights as coarse-grained units representing whole-task knowledge. As a result, irrelevant knowledge from previous tasks may also be transferred, potentially leading to interference and degraded performance on the current task. 
However, existing approaches often treat episodic experiences and insights as coarse-grained units representing whole-task knowledge. As a result, knowledge from previous tasks may include irrelevant subgoals, which can distract reasoning and degrade performance on new tasks. 
Building on the earlier example, suppose the LLM-based agent has previously learned the task ``\textit{cleaning a pan and placing it on the countertop}'' and now faces a new task ``\textit{cooling lettuce and placing it on the countertop}''.
When storing the whole-task knowledge in the memory, the agent may retrieve the entire memory unit of the previous task.
This coarse-grained memory unit may include knowledge associated to the irrelevant subgoal ``\textit{cleaning a pan}'', which can divert attention from the reusable placement subgoal ``\textit{placing it on the countertop}'', thus increasing cognitive overhead and hindering performance~\cite{xiong2025memory}.
This example highlights the importance of transferring only minimally task-relevant fragments of knowledge in multi-task settings.

% Our work addresses this fundamental limitation by developing hierarchical memory architectures and introducing HHR that enable fine-grained and modular knowledge transfer in multi-task settings.
% In this paper, we propose a novel hierarchical memory architecture that fills a critical gap in enabling fine-grained knowledge transfer across tasks in multi-task scenarios.
To address this issue, we propose a hierarchical memory framework that is structured into a high-level planning memory and a low-level execution memory. 
To construct and refine this memory, we introduce a Hierarchical Hindsight Reflection ($H^{2}R$) mechanism that distills agent–environment interactions from past tasks into structured, semantically meaningful memory representations. 
At test time, when dealing with new tasks, the agent can selectively retrieve high-level memories for subgoal planning and low-level memories for action execution, which allows for more targeted and efficient knowledge reuse.
This architecture leads to improved robustness and efficiency in generalization across diverse multi-task scenarios. To validate the efficacy of the framework, we conduct experiments across two benchmarks, including AlfWorld and PDDLGame. 
Experimental results demonstrate that our framework enables more efficient knowledge transfer and outperforms existing baselines like Expel.

The remainder of this paper is organized as follows: Section~\ref{sec-2} reviews related works, Section~\ref{sec-3} formalizes the problem setting, Section~\ref{sec-4} presents the proposed $H^2R$ mechanism, Section~\ref{sec-5} reports experimental results and analysis, and Section~\ref{sec-6} concludes with future directions.

\section{Related Works}
\label{sec-2}
\subsection{LLMs in Decision Making}
When an LLM acts as a decision-making agent, it conditions its textual output on the input prompt's contextual state representation, which is a composite encoding of environmental observations, objectives, and relevant information~\cite{wang2024survey}. 
This foundational capability is further significantly enhanced by structured reasoning frameworks such as Chain-of-Thought (CoT)~\cite{wei2022chain}, Reasoning-Acting (ReAct)~\cite{yao2023react}, Retrieval-Augmented Generation (RAG)~\cite{gao2023retrieval}, etc.
Recent advances have expanded the decision-making and reasoning abilities of LLM-based agents across diverse domains, including embodied robotics~\cite{kim2024survey}, clinical diagnostics~\cite{wang2025survey}, quantitative finance~\cite{nie2024survey}, and others~\cite{hu2024survey,rane2023contribution}.

\subsection{Retrieval-Augmented Generation}
Retrieval-Augmented Generation (RAG) enhances LLMs by integrating external evidence such as knowledge bases or the web~\cite{gao2023retrieval}.
However, its reliance on such resources hinders deployment in scenarios where they are inaccessible or costly, a common situation in agent–environment interactions.
Inspired by RAG, we shift the retrieval of external existing knowledge inward to the retrieval of internal generated memories, i.e., deriving knowledge directly from agent trajectories and reflections to enable context-sensitive reuse of task-relevant experience and improve decision making in new tasks.

\subsection{Experiential Learning for LLM Agents}
Experiential learning encodes agent–environment interactions into compact representations and reflective summaries, which can be selectively retrieved to guide actions in new tasks.
Recent approaches have instantiated this paradigm in diverse ways: Reflexion~\cite{shinn2023reflexion} encodes feedback into textual directives stored in episodic memory for self-corrective learning; Voyager~\cite{wang2023voyager} builds a skill library by synthesizing reusable code primitives from trajectories; and methods such as ExpeL~\cite{zhao2024expel}, RAP~\cite{kagaya2024rap}, and Skill Set Optimization~\cite{nottingham2024skill} extract insights or trajectories from prior tasks to enhance decision making.
Further work incorporates causal analysis to ground responses and reduce hallucination~\cite{sun2024enhancing,chari2025mindstores}.
Nonetheless, most existing works treat episodic memories as coarse-grained units, risking irrelevant knowledge transfer that can impair performance, whereas recent findings highlight the need for selective and modular memory reuse~\cite{xiong2025memory}.

\section{{Problem Statement}}
\label{sec-3}
We consider an autonomous LLM agent embedded within a discrete interactive environment to sequentially execute predefined tasks. 
Similar to Reinforcement Learning (RL)~\cite{sutton1999reinforcement}, such interaction process could be modeled as a Partially Observable Markov Decision Process (POMDP), which provides a mathematical framework for sequential decision making under uncertainty, formally defined by the tuple $\langle \mathcal{S}, \mathcal{A}, \mathcal{O}, T, \Omega, R, \gamma\rangle$. Under this formulation, an agent receives only ambiguous observations $o_t \in \mathcal{O}$ generated through the observation function $\Omega(s, a) = \Pr(o | s, a)$. 
The historical interaction trajectory at timestep $t$ is formalized as $h_t = (o_0, \dots, a_t, o_{t})$, where $(a_t, o_{t})$ represents action-observation pairs at step $t$. 
Particularly, at the final timestep $T$, $h_{T}$ captures the complete sequence of interactions, denoted as the full trajectory $\tau$.
% Together with the goal $g$ and relevant knowledge $\mathcal{K}$, these elements form the structured input prompt $p_t = (h_t, g, \mathcal{K})$.
The structured input prompt for the agent at timestep $t$ is then formed from the current trajectory $h_t$, the task goal $g$, and relevant knowledge $\mathcal{K}$, i.e., $p_t = (h_t, g, \mathcal{K})$. 
The policy $\pi_\theta$, instantiated by an LLM with parameters $\theta$, generates the subsequent action $a_{t+1}$ via policy mapping $\pi_\theta: p_t \mapsto a_{t+1}$.

In this work, we investigate efficient knowledge utilization strategies to empower future task decision-making performance. 
Focusing on multi-task settings, we consider an agent operating in a multi-task environment with dataset $\mathcal{D}= \{ \mathcal{D}_1, \dots, \mathcal{D}_n \}$, where tasks are split into a training set $\mathcal{D}_{\text{train}}$ and a testing set $\mathcal{D}_{\text{test}}$. The agent acquires experiential knowledge through interactions with tasks in $\mathcal{D}_{\text{train}}$ and is then required to leverage this knowledge when solving unseen tasks in $\mathcal{D}_{\text{test}}$. The core challenge lies in enabling efficient knowledge transfer when encountering new tasks $\mathcal{D}_{\text{new}} \in \mathcal{D}_{\text{test}}$.

\begin{figure*}[htbp]
  \centering
  \includegraphics[width=1.0\textwidth]{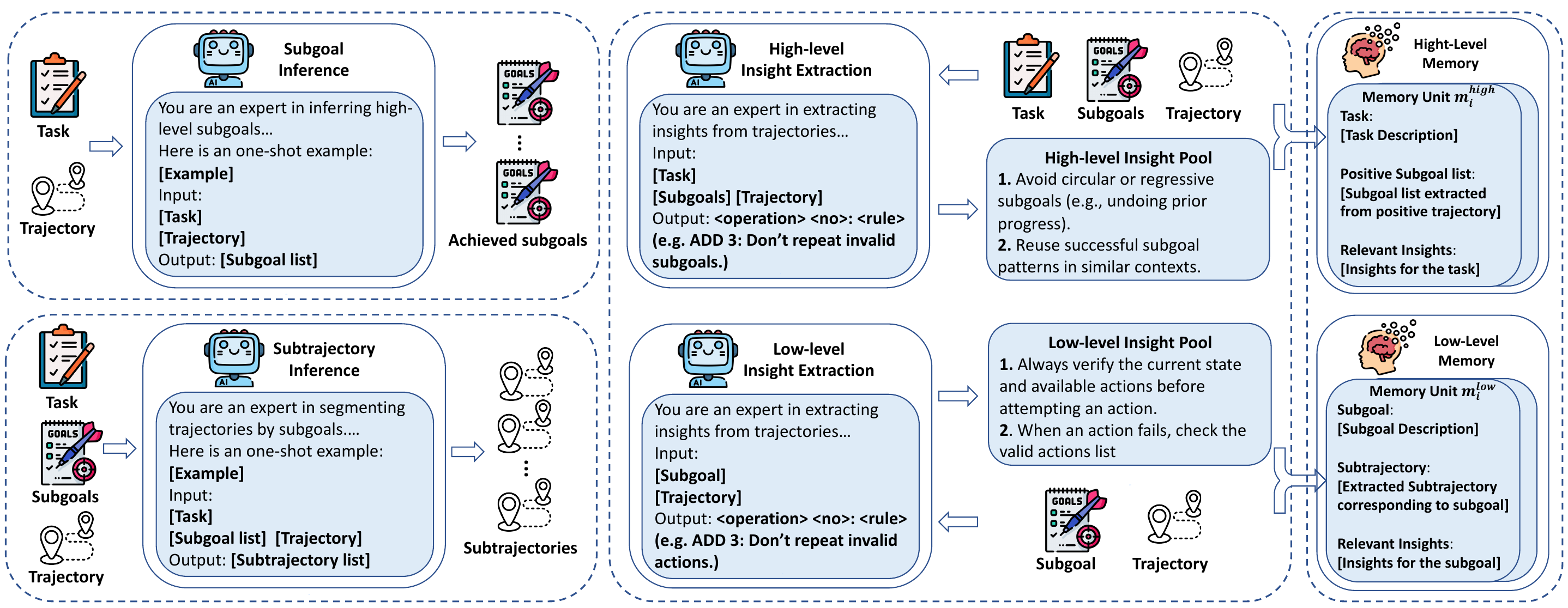}
  \caption{Overview of \textit{Hierarchical Hindsight Reflection ($H^2R$)} framework, which consists of four key processes: (1) \textbf{Subgoal Inference}, which decomposes tasks into achieved subgoals given tasks and corresponding task trajectories; (2) \textbf{Subtrajectory Inference}, which segments trajectories by subgoals to extract subtrajectory sequences; (3) \textbf{Insight Extraction}, performed at both high-level (from tasks, subgoals, and trajectories) and low-level (from individual subgoals and their trajectories) to derive reusable and beneficial rules; and (4) \textbf{Memory Organization}, where relevant insights are attached to corresponding memory units. This architecture enables efficient knowledge transfer through level-specific retrieval mechanisms that effectively decouple high-level planning from low-level execution in multi-task scenarios.}
  \label{fig:example}
\end{figure*}

\section{Hierarchical Hindsight Reflection}
\label{sec-4}
To tackle the challenge of efficient knowledge transfer in multi-task scenarios, we begin by proposing a hierarchical memory architecture that organizes knowledge into high-level and low-level components. To construct and refine these memories, we introduce $H^2R$, which distills task experiences into reusable memory units at different levels of granularity. Leveraging these memories, the agent can then plan and act in a hierarchical manner, where a high-level Planner decomposes tasks into subgoals and a low-level Executor carries out these subgoals through atomic actions.
This decoupling enables the agent to retrieve and reuse task-relevant knowledge at the appropriate granularity, thereby improving decision making in multi-task settings. 
Fig.~\ref{fig:example} provides an overview of the core reflection workflow underlying the proposed architecture.

\subsection{The Hierarchical Memory Architecture}
Our hierarchical memory architecture consists of a high-level memory component $\mathcal{M}_{\mathrm{high}}$ that captures abstract task structures and a low-level memory component $\mathcal{M}_{\mathrm{low}}$ that encodes detailed interaction mechanics for executing subgoals. 
This architectural decoupling enables each memory unit in the low-level memory component \(\mathcal{M}_{\mathrm{low}}\) to specialize in its own atomic subgoal, thereby mitigating interference from irrelevant knowledge inherent to naive memory designs.

Specifically, both memory components consist of structured memory units that support efficient retrieval of relevant knowledge. As shown in Fig.~\ref{fig:example}, each high-level memory unit in $\mathcal{M}_{\mathrm{high}}$ consists of the task description $\mathcal{X}$, the sequence of realized subgoals during execution $\mathcal{G}$, and the planning insights $\mathcal{I}_{\mathrm{plan}}$.  
Similarly, each low-level memory unit in $\mathcal{M}_{\mathrm{low}}$ contains a specific subgoal $g$, the corresponding detailed interaction trajectory $\tau$, and the fine-grained execution insights $\mathcal{I}_{\mathrm{low}}$.
%Each component of the memory unit is described using natural language. 
Unlike RAG systems, these memory units are derived from agent–environment interactions and hindsight reflection processes, rather than being provided a priori.
When confronting tasks with analogous inner structure, relevant memory units, which encode experiential patterns along with reflective insights, will be activated to  facilitate cross-task knowledge transfer and enhance decision-making performance.

\subsection{The Construction of Memory Units}
To gather experiences for knowledge extraction and further memorial retrieval, we first perform tasks from the training set $\mathcal{D}_{\text{train}}$.
During task execution, $H^2R$ employs the Planner to generate subgoals and the Executor to carry out these subgoals through atomic actions.
However, the agent, particularly its high-level Planner, has limited knowledge of the underlying task inner structure and may therefore generate inappropriate subgoals that contribute little to or even hinder task progress. In such cases, even if the Executor correctly accomplishes the given subgoal, the overall task may still fail to proceed. To address this problem, during training (or collecting experiences), the high-level Planner is constrained to output the current task directly instead of generating any subgoals. 
To improve the likelihood of success in the next attempt of the same task, Reflexion~\cite{shinn2023reflexion} is employed to analyze the failed trajectories, following the experience-gathering strategy of Expel~\cite{zhao2024expel}.

Based on the collected trajectories, $H^2R$ analyzes task executions to extract and refine structured memory units in the high- and low-level components, capturing abstract strategies for task planning and detailed patterns for subgoal execution. Specifically, the process performs high-level reflection to construct high-level memory units and low-level reflection to create low-level memory units, with each reflection capturing knowledge at its respective level of granularity.

\begin{figure*}[htbp]
  \centering
  \includegraphics[width=1.0\textwidth]{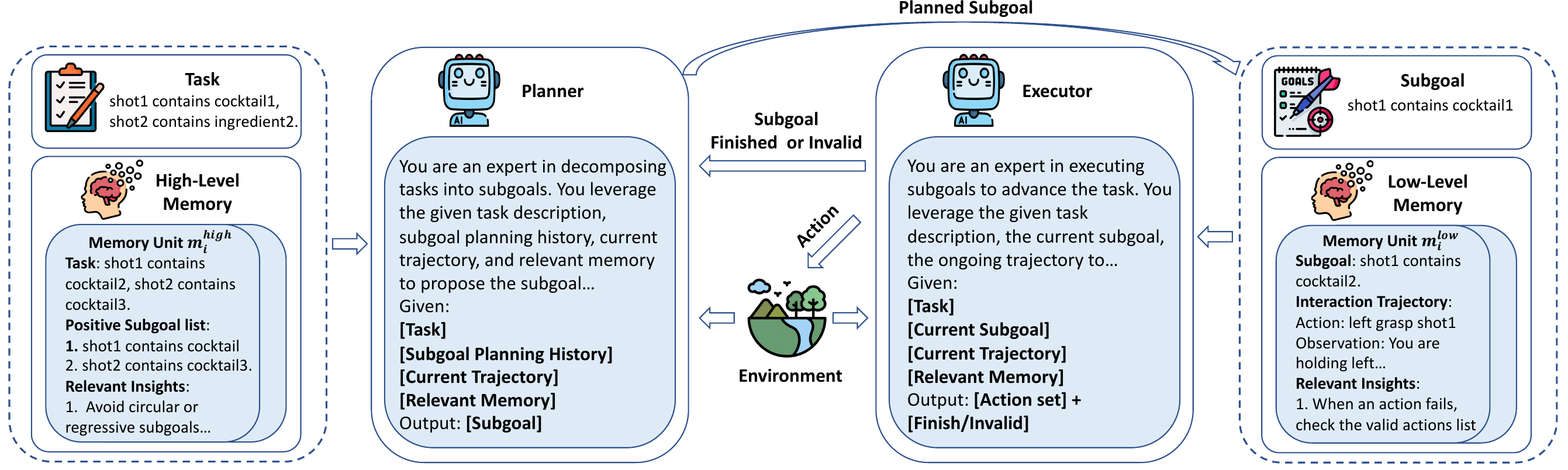}
  \caption{Overview of utilization of memory components. The system comprises three core components: (1) \textbf{Memory Module} featuring two specialized components: (a) the \textit{High-Level Memory Component} containing memory units ($m_i^{\text{high}}$) that store task description, subgoal sequence, and planning insights and (b) the \textit{Low-Level Memory Component} containing memory units ($m_i^{\text{low}}$) that store subgoal description, execution trajectory, and execution insights. For any given task, relevant memory units from both components are retrieved to inform decision making. (2) \textbf{Planner} that decomposes tasks into subgoals using task descriptions, planning history, current trajectories, and retrieved high-level memory, outputting structured subgoals like ``\textit{shot1 contains cocktail2}''. (3) \textbf{Executor} that translates subgoals into actionable steps using task context, current subgoals, ongoing trajectories, and retrieved low-level memory, generating action (e.g., ``\textit{left grasp shot1}'') or termination signals.}
  \label{fig:test-time}
\end{figure*}

\subsubsection{\textbf{High-level Reflection}}
This process is designed to extract high-level memory units through reflective processing. 
Given a set of trajectories, we generate high-level memory units, each consisting of three components: the task description $\mathcal{X}$, the sequence of realized subgoals during execution $\mathcal{G}$, and the planning insights $\mathcal{I}_{\mathrm{plan}}$. The high-level reflection process consists of two steps: subgoal inference and insight extraction.

\textbf{Subgoal Inference.} Subgoal inference is introduced to extract the sequence of subgoals realized from a given task and the interaction trajectory. The intuition behind this process is that, in order to generate useful knowledge about subgoal planning, we must first assess the quality of the subgoals proposed by the Planner. However, a single interaction trajectory does not directly reveal the corresponding subgoal sequence, nor does it guarantee that the Executor has accurately executed these subgoals. Therefore, we infer the subgoal sequence through a hindsight reflection process and assume that the Executor successfully executes them, given that the inferred subgoals are grounded in the actions the agent actually performs. Formally, for a given task $\mathcal{X}^i$ and the corresponding interaction trajectory $\tau^i$, reflection is conducted to infer subgoal sequence by prompting an LLM: 
\begin{equation}
    \mathcal{G}^i \leftarrow \mathcal{F}_{subgoal}(\mathcal{X}^i, \tau^i),
\end{equation}
with $\mathcal{G}^i = \{g^i_1, \ldots, g^i_k\}$ denoting the inferred sequence of subgoals.

\textbf{Insight Extraction}. 
Based on the inferred subgoals, high-level insights are extracted and maintained in a fixed-size set $\mathcal{I}_{high}$, using the mechanism proposed in ExpeL~\cite{zhao2024expel}.
To update the set, contrastive reflection is applied to analyze strategies that lead to success while identifying potential causes of failure.
This reflection process is implemented by prompting an LLM to perform four types of operations on $\mathcal{I}_{high}$: \texttt{add} (introduce a new insight), \texttt{modify} (refine an existing insight), \texttt{upvote} (increase the importance of an insight), and \texttt{downvote} (decrease the importance of an insight). 
Formally, this contrastive reflection and insight updating process can be represented as a function $\mathcal{F}_{high}$, which takes the current task $\mathcal{X}^i$, its successful trajectory $\tau_{+}^i$ and failed trajectory $\tau_{-}^i$ with subgoal sequences $\mathcal{G}_{+}^i$ and $\mathcal{G}_{-}^i$, and the existing set of insights $\mathcal{I}_{high}$ as input, and outputs the updated set of high-level insights:

\begin{equation}
\label{eq:extract_insight}
    \mathcal{I}_{high} \leftarrow \mathcal{F}_{high}(\mathcal{X}^i,\tau_{+}^i,\tau_{-}^i, \mathcal{G}_{+}^i,\mathcal{G}_{-}^i,\mathcal{I}_{high}).
\end{equation}

% Each insight is associated with an importance counter, which is incremented upon \texttt{upvote} or \texttt{modify} and decremented upon \texttt{downvote}. Insights with zero importance are pruned. This mechanism ensures that $\mathcal{I}_{high}$ remains concise, relevant, and high-quality across diverse tasks. 

% To manage the resulting insight set $\mathcal{I}_{high}$, we adopt the insight maintenance mechanism from the Expel framework, which allows the LLM to perform operations including \texttt{add}, \texttt{modify}, \texttt{upvote}, and \texttt{downvote}, ensuring the insight set $\mathcal{I}_{high}$ maintains high quality across tasks xxxx:
% \begin{equation}
% \label{eq:extract_insight}
%     \mathcal{I}_{high} \leftarrow \mathcal{F}_{high}(\mathcal{X}^i,\tau_{+}^i,\tau_{-}^i, \mathcal{G}_{+}^i,\mathcal{G}_{-}^i,\mathcal{I}_{high}).
% \end{equation}

Once the high-level reflection process is finished and all insights have been generated, a memory unit is constructed for each task, which is formed by combining three key components: the task description itself $\mathcal{X}^i$, its corresponding successful subgoal sequence $\mathcal{G}_{+}^i$, and the relevant insights $\mathcal{I}_{high}^{i}$. 
% The grounding of task-relevant insights $\mathcal{I}_{high}^{i}$ is efficiently handled by an LLM, denoted as $F_{ground}$.
An LLM-based grounding function $F_{ground}$ selects the task-relevant high-level insights $\mathcal{I}_{high}^{i}$ from the full set $\mathcal{I}_{high}$ by evaluating the relevance of each insight in $\mathcal{I}_{high}$ to the current task $\mathcal{X}^i$.
The resulting high-level memory unit $\{\mathcal{X}^i,\mathcal{G}^i_+, \mathcal{I}_{high}^i\}$ thus encapsulates the essential abstractions and high-level insights derived from raw trajectories. 
These abstractions form the foundation for subsequent low-level reflection.

\subsubsection{\textbf{Low-level Reflection}}
This process is responsible for extracting low-level memory units, which capture fine-grained execution details and insights about action-level patterns.
To guarantee that each low-level memory unit captures information specific to a single subgoal, the sub-trajectory corresponding to each subgoal is first extracted and subsequently analyzed to derive low-level insights.  

To ensure the reliability of the low-level memory units, only subgoals inferred from successful trajectories are utilized. Formally, given a task $\mathcal{X}^i$, a successful trajectory $\tau_{+}^i$ and its corresponding sequence of implemented subgoals $\mathcal{G}_{+}^i$, an LLM extracts the associated sub-trajectories as follows:
\begin{equation}
    \mathcal{T} \leftarrow \mathcal{F}_{trajectory}(\mathcal{X}^i,\tau_{+}^i,\mathcal{G}_{+}^i),
\end{equation}
with $\mathcal{T} = \{\tau_{+,1}^i, \ldots, \tau_{+,k}^i\}$ denoting the extracted sub-trajectories corresponding to the given subgoals. Reflection is then performed based on each subgoal $g^i$, its corresponding trajectory $\tau_+^i$, and a failed trajectory $\tau_{-}^i$ of the same task. 
% Low-level reflection does not follow a strictly contrastive form, as $g^i$ may still be accomplished within a failed trajectory.  
As in high-level reflection, the insight extraction mechanism is also employed and formulated for low-level reflection as follows:
\begin{equation}
 \mathcal{I}_{low} \leftarrow  \mathcal{F}_{low}(g^i,\tau^{i}_+,\tau_{-}^i,\mathcal{I}_{low}).
\end{equation}
% We constrain the scope of reflection to a specific subgoal $g^i$, thereby extracting fine-grained insights tailored to it. 
Then, in a manner analogous to the construction of high-level memory units, we retrieve relevant insights for a specific subgoal $g^i$ using an LLM and formulate them into a low-level memory unit $\{g^i, \tau^{i}_+,\mathcal{I}_{low}^i\}$. Since the content of low-level reflection is grounded in that of high-level reflection, we refer to our approach as Hierarchical Hindsight Reflection ($H^2R$). The overall workflow of $H^2R$ is illustrated in Algorithm~\ref{alg:algorithm}.

\subsection{The Utilization of Memory Units}

After completing the overall reflection process, the extracted high-level and low-level memory units can be applied to new tasks.
Specifically, as shown in Fig.~\ref{fig:test-time}, the Planner relies on high-level memories, while the Executor draws exclusively on low-level memories.
Rooted in the principle of hierarchical task decomposition, the Planner formally maps natural language task specifications into structured intermediate subgoals and enhances its subgoal planning for the current task $\mathcal{X}$ by utilizing relevant memories.
To retrieve such memories,  vector embeddings of the current task description and stored task descriptions in the high-level memory component are computed using a pretrained sentence encoder $e$. By measuring cosine similarity between both embeddings, we retrieve the top-$k$ most relevant memory units:
\begin{equation}
\mathcal{M}_{\mathrm{high}}^{\mathrm{relevant}} = \underset{m^i_{high} \in \mathcal{M}_{high}}{\text{top-}k} [sim(\mathcal{X},\mathcal{X}^i)].
\end{equation}

\begin{algorithm}[t]
\caption{Hierarchical Hindsight Reflection}
\label{alg:algorithm}
\textbf{Input}: Collected trajectories $\mathcal{T}$, subgoal inference module $\mathcal{F}_{subgoal}$, high-level reflection module $\mathcal{F}_{high}$, sub-trajectory partition module $\mathcal{F}_{trajectory}$, low-level reflection module $\mathcal{F}_{low}$ 
\begin{algorithmic}[1] %[1] enables line numbers
\STATE Initialize the high-level memory component $\mathcal{M}_{\mathrm{high}}$, the low-level memory component $\mathcal{M}_{\mathrm{low}}$, the high-level insight set $\mathcal{I}_{high}$ and the low-level insight set $\mathcal{I}_{low}$. 
\FOR{Each pair $\mathcal{X}^i, \tau^i_{+},\tau^i_{-}$ $\in$ $\mathcal{T}$}
\STATE \textcolor{blue}{\# extract subgoal sequences of $\tau^i_{+},\tau^i_{-}$}
\STATE $\mathcal{G}^i_{+} \leftarrow \mathcal{F}_{subgoal}(\mathcal{X}^i,\tau^i_{+})$
\STATE $\mathcal{G}^i_{-} \leftarrow \mathcal{F}_{subgoal}(\mathcal{X}^i,\tau^i_{-})$
\STATE \textcolor{blue}{\# update high-level insights about planning}
\STATE $\mathcal{F}_{high}(\mathcal{X}^i,\tau_{+}^i,\tau_{-}^i, \mathcal{G}_{+}^i,\mathcal{G}_{-}^i,\mathcal{I}_{high})$
\STATE add $\{\mathcal{X}^i,\mathcal{G}^i_{+}, \emptyset \}$ to $\mathcal{M}_{\mathrm{high}}$
\STATE \textcolor{blue}{\# partition the positive trajectory into sub-trajectories}
\STATE $\mathcal{T}_{sub}^i \leftarrow \mathcal{F}_{trajectory}(\tau_{+}^i,\mathcal{G}_{+}^i),$
\FOR{Each subgoal $g^i_j \in \mathcal{G}^i_{+}$}
\STATE \textcolor{blue}{\# update low-level insights about execution}
\STATE $ \mathcal{F}_{low} (g^i_j,\tau_{+},\tau_{-},\mathcal{I}_{low})$
\STATE add $\{g^i_j,\tau_+^i, \emptyset\}$ to $\mathcal{M}_{\mathrm{low}}$
\ENDFOR
\ENDFOR
\STATE \textcolor{blue}{\# attach relevant insights to memory units}
\FOR{Each high-level memory unit $m_{\mathrm{high}}^i$ in $\mathcal{M}_{\mathrm{high}}$}
\STATE $\mathcal{I}_{high}^i \leftarrow F_{ground}(m_{\mathrm{high}}^i,\mathcal{I}_{high})$
\STATE replace $m_{\mathrm{high}}^i$ by $\{\mathcal{X}^i,\mathcal{G}^i_{+}, \mathcal{I}_{high}^i \}$
\ENDFOR
\FOR{Each high-level memory unit $m_{\mathrm{low}}^i$ in $\mathcal{M}_{\mathrm{low}}$}
\STATE $\mathcal{I}_{low}^i \leftarrow F_{ground}(m_{\mathrm{low}}^i,\mathcal{I}_{low})$
\STATE replace $m_{\mathrm{low}}^i$ by $\{g_i,\mathcal{\tau}^i_{+}, \mathcal{I}_{low}^i \}$
\ENDFOR
\STATE \textbf{return} $\mathcal{M}_{\mathrm{high}},\mathcal{M}_{\mathrm{low}}$
\end{algorithmic}
\end{algorithm}

Functioning as the action grounding module, the Executor grounds the textual subgoal $g$ from the Planner into executable atomic  action selected from a predefined set \(\mathcal{A} = \{a_1, \ldots, a_K\} \cup \{a_{\mathrm{+}},a_{\mathrm{-}}\}\), where \( a_1 \) to \( a_K \) are domain-specific atomic actions, \( a_{\mathrm{+}} \) is a primitive indicating subgoal completion and \( a_{\mathrm{-}} \) signals that the subgoal is invalid.
When the Executor determines that the current subgoal $g$ has been completed or is unachievable, it outputs \( a_{\mathrm{+}} \) or \( a_{\mathrm{-}} \) to trigger the Planner to replan a new subgoal. Similarly, the top-$k$ most relevant low-level memory units are retrieved by computing the semantic similarity between the current subgoal description and the subgoal descriptions stored within these units:
\begin{equation}
\mathcal{M}_{\mathrm{low}}^{\mathrm{relevant}} = \underset{m^i_{low} \in \mathcal{M}_{low}}{\text{top-}k} [sim(g,g^i)].
\end{equation}
% where the $sim$ function is realized as Formula~\ref{eq:cos}. 
In practice, FAISS~\cite{douze2024faiss} can be employed for efficient high-dimensional similarity search, and retrieval performance can be further enhanced through existing techniques such as reranking~\cite{glass2022re2g} and rewriting~\cite{liu2024query}. 

By organizing memory hierarchically, $H^2R$ allows LLM-based agents to selectively access only the knowledge necessary for the current context, reducing interference from irrelevant experiences and enabling more robust and efficient decision making in multi-task scenarios. 

\section{Experiments}
\label{sec-5}
We conduct experiments to demonstrate the effectiveness of $H^2R$ across diverse multi-task environments. The experimental design addresses two key research questions: (1) How does hierarchical memory organization compare to existing memory architectures in multi-task scenarios? and (2) What are the individual contributions of different memory components in our framework?

\subsection{Experimental Setup}
Experiments are conducted on AlfWorld (a text-based household environment)~\cite{shridhar2020alfworld} and PDDLGame (a strategic game environment)~\cite{chang2024agentboard}, with comparisons against two representative baselines: (1) ReAct~\cite{yao2023react}, the foundational reasoning-acting paradigm without memory mechanisms; and (2) ExpeL~\cite{zhao2024expel}, which extracts insights from successful and failed trajectories.
The datasets of both benchmarks are split in half for training and evaluation, with a maximum of 30 steps per episode in AlfWorld and 40 steps in PDDLGame.
Six task types from ALFWorld (\textit{pick\_and\_place}, \textit{pick\_clean\_then\_place}, \textit{pick\_heat\_then\_place}, \textit{pick\_cool\_then\_place}, \textit{look\_at\_obj}, \textit{pick\_two\_obj}) and three types from PDDLGame (\textit{barman}, \textit{gripper}, \textit{tyreworld}) are included in the datasets.
All agent components, including reflection, planning, and execution, are implemented using Qwen3-235B-A22B-Instruct-2507, and semantic similarity for memory retrieval is computed using Qwen3-Embedding-0.6B. Methods are evaluated on three held-out test episodes per environment, with results averaged across three independent runs to ensure statistical reliability.

\begin{table}[H]
\caption{Comparison Results}
\centering
\begin{tabular}{|p{0.1\textwidth}|p{0.15\textwidth}|p{0.15\textwidth}|} 
\hline
\makecell{\multirow{2}{*}{\textbf{Algorithms}}} & \multicolumn{2}{|c|}{\textbf{Success Rate on Benchmarks(\%)}} \\
\cline{2-3} 
\centering
 & \textbf{\makecell{AlfWorld}}& \textbf{\makecell{PDDLGame}} \\
\hline
\makecell{ReAct} & \makecell{46.3} & \makecell{66.7} \\
\hline
\makecell{Expel} & \makecell{72.4} & \makecell{72.2} \\
\hline
\makecell{$H^2R$} & \makecell{75.9} & \makecell{80.5}  \\
\hline
\end{tabular}
\label{tab1}
\end{table}

\label{comparison-results}
\subsection{Comparison Results}
Table 1 summarizes the performance comparison across both benchmark environments. As can be seen, $H^2R$ outperforms all baselines, with success rates of 75.9\% in AlfWorld and 80.5\% in PDDLGame, representing relative improvements of 3.5\% and 8.3\% over the baseline ExpeL. The performance gains validate our core hypothesis that hierarchical memory organization enables more effective knowledge transfer in multi-task settings. Notably, the improvements are most pronounced in PDDLGame, which involves more complex hierarchical planning requirements, demonstrating the particular strength of our approach in complex decision-making scenarios.
These observations highlight the fundamental advantage of decoupling high-level planning knowledge from low-level execution patterns, enabling more precise and interference-free knowledge transfer across diverse multi-task environments.

\subsection{Ablation Studies}
To evaluate the individual contribution of each component in our hierarchical framework, we conduct ablation experiments by examining the contribution of different levels in our $H^2R$ mechanism by selectively removing high-level or low-level memory units. When eliminating high-level memory units, the system cannot extract task-level insights and subgoal sequences, forcing it to operate without strategic planning knowledge. This results in performance degradation of 27.7\% in PDDLGame. Conversely, removing low-level memories prevents the utilization of execution insights and subgoal-specific patterns, leading to 19.4\% performance drops. These results demonstrate that both reflection levels are essential for the comprehensive knowledge extraction.

\begin{table}[t]
\caption{Ablation Studies}
\centering
\begin{tabular}{|p{0.20\textwidth}|p{0.14\textwidth}|} 
\hline
\makecell{\textbf{Algorithms}} & \makecell{\textbf{Success Rate (\%)}}\\
\hline
\makecell{$H^2R$} & \makecell{80.5} \\
\hline
\makecell{$H^2R$ w/o high-level memories} & \makecell{52.8} \\
\hline
\makecell{$H^2R$ w/o low-level memories} & \makecell{61.1} \\
\hline
\end{tabular}
\label{tab1}
\end{table}

\section{Conclusion}
\label{sec-6}
In this work, we propose a novel hierarchical memory architecture that decouples high-level planning memory from low-level execution memory, enabling fine-grained knowledge transfer in multi-task scenarios. By distilling reusable and hierarchical knowledge from past agent-environment interactions and performing retrieval separately at each
memory level, our framework selectively reuses specialized knowledge relevant to the current context. Experimental results across AlfWorld and PDDLGame demonstrate that our framework improves generalization and decision-making performance, outperforming strong baselines such as ReAct and ExpeL. 
Future work will extend $H^2R$ to more complex and dynamic environments, while supporting multi-agent scenarios to facilitate collaborative decision making and knowledge sharing.

\bibliographystyle{unsrt}
% \bibliography{refs}

\vspace{12pt}

\end{document}